\documentclass[runningheads,a4paper,twoclumns]{llncs}\usepackage[]{graphicx}\usepackage[]{color}
\makeatletter
\def\maxwidth{ %
  \ifdim\Gin@nat@width>\linewidth
    \linewidth
  \else
    \Gin@nat@width
  \fi
}
\makeatother

\definecolor{fgcolor}{rgb}{0.345, 0.345, 0.345}

\usepackage{framed}
\makeatletter
 {\par\unskip\endMakeFramed%
 \at@end@of@kframe}
\makeatother

\definecolor{shadecolor}{rgb}{.97, .97, .97}
\definecolor{messagecolor}{rgb}{0, 0, 0}
\definecolor{warningcolor}{rgb}{1, 0, 1}
\definecolor{errorcolor}{rgb}{1, 0, 0}
\newenvironment{knitrout}{}{} 

\usepackage{alltt}

\usepackage{graphicx}
\usepackage{bm}
\usepackage{amsmath}
\usepackage{dsfont}
\usepackage{booktabs}
\usepackage{algorithm}
\usepackage{csquotes}
\usepackage{hyperref}
\usepackage{subcaption}
\usepackage[utf8]{inputenc}
\usepackage[T1]{fontenc}
\usepackage[figuresleft]{rotating}

\usepackage{url}
\urldef{\mailsa}\path|{firstname.lastname}@stat.uni-muenchen.de|
\urldef{\mailsb}\path|tobias.hepp@uk-erlangen.de|

\newcommand{\ms}{\ensuremath{m_\text{stop}}}

\IfFileExists{upquote.sty}{\usepackage{upquote}}{}
\begin{document}

\mainmatter


\title{Probing for sparse and fast variable selection with model-based boosting}
\titlerunning{Probing for sparse and fast variable selection with model-based boosting}
\toctitle{Probing for sparse and fast variable selection with model-based boosting}

\author{Janek~Thomas\inst{1,*}\and Tobias~Hepp\inst{2,*} \and Andreas~Mayr\inst{2}\and Bernd~Bischl\inst{1}}
\authorrunning{Thomas J., Hepp T. et al.}
\tocauthor{Janek~Thomas, Tobias~Hepp, Bernd~Bischl}

\institute{
  LMU M{\"u}nchen, Department of Statistics,\\
  \mailsa\\
  \and
  FAU Erlangen-Nürnberg, Department of Medical Informatics, Biometry and Epidemiology,\\
  \mailsb
  \\\vspace{1ex}
  $\ast$: Authors with equal contributions
}

\maketitle

\begin{abstract}
We present a new variable selection method based on model-based gradient boosting and randomly permuted variables.
Model-based boosting is a tool to fit a statistical model while performing variable selection at the same time.
A drawback of the fitting lies in the need of multiple model fits on slightly altered data (e.g. cross-validation or bootstrap) to find the optimal number of boosting iterations and prevent overfitting.
In our proposed approach, we augment the data set with randomly permuted versions of the true variables, so called shadow variables, and stop the step-wise fitting as soon as such a variable would be added to the model.
This allows variable selection in a single fit of the model without requiring further parameter tuning.
We show that our probing approach can compete with state-of-the-art selection methods like stability selection in a high-dimensional classification benchmark and apply it on gene expression data for the estimation of riboflavin production of Bacillus subtilis.
\end{abstract}
\section{Introduction}
\label{sec:intro}

At the latest since the emergence of genomic and proteomic data, where the number of available variables \emph{p} is possibly far higher than the sample size \emph{n}, high-dimensional data analysis becomes increasingly important in biomedical research \cite{romero2006use,clarke2008properties,mallick2010proteomics,bermingham2015application}.
Since common statistical regression methods like ordinary least squares are unable to estimate model coefficients in these settings due to singularity of the covariance matrix, 
varying strategies have been proposed to select only truly influential, i.e., informative variables and discard those without impact on the outcome.

By enforcing sparsity in the true coefficient vector
, regularized regression approaches like the \emph{lasso} \cite{tibshirani96lasso}, \emph{least angle regression} \cite{efron2004lars}, \emph{elastic net} \cite{zou2005elastic} and \emph{gradient boosting} algorithms \cite{friedman2000boosting,buehlmann07boosting} perform variable selection directly in the model fitting process.
This selection is controlled by tuning hyperparameters that define the degree of penalization. 
While these hyperparameters are commonly determined using resampling strategies like cross-validation, bootstrapping and similar methods, the focus on minimizing the prediction error often results in the selection of many noninformative variables \cite{meinshausen2006high,leng2006note}.

One approach to address this problem is \emph{stability selection}~\cite{meinshausen2010stab,shah2013variable}, a method that combines variable selection with repeated subsampling of the data to evaluate selection frequencies of variables.
While stability selection can considerably improve the performance of several variable selection methods including regularized regression models in high dimensional settings~\cite{meinshausen2010stab,hofner2014controlling}, its application depends on additional hyperparameters.
Although recommendations for reasonable values exist \cite{meinshausen2010stab,hofner2014controlling}, proper specification of these parameters is not straightforward in practice as the optimal configuration would require a priori knowledge about the number of informative variables.
Another potential drawback is that stability selection increases the computational demand, which can be problematic in high-dimensional settings if the computational complexity of the used selection technique scales superlinearly with the number of predictor variables.

In this paper, we propose a new method to determine the optimal number of iterations in model-based boosting for variable selection inspired by \emph{probing}, a method frequently used in related areas of machine learning research \cite{guyon2003introduction,bi2003sparse,wu2007controlling} and the analysis of microarrays \cite{tusher2001significance}.
The general notion of probing involves the artificial inflation of the data with random noise variables, so-called \emph{probes} or \emph{shadow variables}.
While this approach is in principle applicable to the lasso or least angle regression as well, it is especially attractive to use with more computationally intensive boosting algorithms, as no resampling is required at all.
Using the first selection of a shadow variable as stopping criterion, the algorithm is applied only once without the need to optimize any hyperparameters in order to extract a set of informative variables from the data, thereby making its application very fast and simple in practice.
Furthermore, simulation studies show that the resulting models in fact tend to be more strictly regularized compared to the ones resulting from cross-validation and contain less uninformative variables.

In Section 2, we provide detailed descriptions of the model-based gradient boosting algorithm as well as stability selection and the new probing approach.
Results of a simulation study comparing the performance of probing to cross-validation and different configurations of stability selection in a binary classification setting are then presented in Section 3 before discussing the application of these methods on data of riboflavin production by Bacillus subtilis~\cite{buhlmann2014high} in Section 4.
Section 5 summarizes our findings and presents an outlook to extensions of the algorithm.

\section{Methods}
\label{sec:methods}

\subsection{Gradient boosting}

Given a learning problem with a data set $D = \{(\bm x^{(i)}, y^{(i)})\}_{i=1,...n}$ sampled i.i.d from a distribution over the joint space $\mathcal{X} \times \mathcal{Y}$, with a $p$-dimensional input space $\mathcal{X} = (\mathcal{X}_1\times \mathcal{X}_2\times...\times \mathcal{X}_p)$ and an output space $\mathcal{Y}$ (e.g., $\mathcal{Y} = \mathds{R}$ for regression and $\mathcal{Y} = \{0,1\}$ for binary classification), the aim is to estimate a function $f(\bm x), \; \mathcal{X} \rightarrow \mathcal{Y}$, that maps elements of the input space to the output space as good as possible. 
Relying on the perspective on boosting as gradient descent in function space, gradient boosting algorithms try to minimize a given loss function $\rho(y^{(i)}, f(\bm x^{(i)})), \; \rho:\mathcal{Y} \times \mathds{R} \rightarrow \mathds{R}$, that measures the discrepancy between a predicted outcome value of $f(\bm x^{(i)})$ and the true $y^{(i)}$. 
Minimizing this discrepancy is achieved by repeatedly fitting weak prediction functions, called \emph{base learners}, to previous mistakes, in order to combine them to a strong ensemble \cite{hastie2001statisticallearning}.
Although early implementations in the context of machine learning focused specifically on the use of regression trees, the concept has been successfully extended to suit the framework of a variety of statistical modelling problems \cite{friedman2000boosting,ridgeway1999state}.
In this model-based approach, the \emph{base learners} $h(\bm x)$ are typically defined by semi-parametric regression functions on $\bm x$ to build an additive model. 
A common simplification is to assume that each base learner $h_j$ is defined on only one component $x_j$ of the input space
\[
f(\bm x)=\beta_0+h_1(x_1)+\dots+h_p(x_p). 
\]
For an overview of the fitting process of model-based boosting see Algorithm~\ref{alg:boosting}.

\begin{algorithm}
\caption{Model-based gradient boosting}
\label{alg:boosting}
Starting at $m=0$ with a constant loss minimal initial value $\hat{f}^{[0]}(\bm x) \equiv c$, the algorithm iteratively updates the predictor with a small fraction of the base learner with the best fit on the negative gradient of the loss function:
\begin{enumerate}
\item Set iteration counter $m:=m+1$. 
\item While $m \leq m_{\text{stop}}$, compute the negative gradient vector of the loss function: 
\[ 
 u^{(i)} = -\frac{\partial\rho(y,f)}{\partial f} \Big|_{f=\hat{f}^{[m-1]}(\bm x^{(i)}), y = y^{(i)}} 
\]
\item Fit every base learner $h_{j}^{[m]}(x_j)$ separately to the negative gradient vector $\bm u$
  \item
    Find $\hat{h}^{[m]}_{j^*}(\boldsymbol{x}_{j^*})$, i.e., the base learner with the best fit:
    \[ j^*= \operatorname*{arg\,min }_{1\le j\le p} \sum_{i=1}^n \left(u^{(i)}-\hat{h}^{[m]}_{j}(x_{j}^{(i)})\right)^2 \]
  \item
    Update the predictor with a small fraction $0\le\nu\le 1$ of this component: \[\hat{f}(\bm x)^{[m]}=\hat{f}(\bm x)^{[m-1]}+\nu\cdot\hat{h}^{[m]}_{j^*}(x_{j^*})\]
\end{enumerate}
\end{algorithm}

The resulting model can be interpreted as a generalized additive model with partial effects for each covariate contained in the additive predictor.
Although the algorithm relies on two hyperparameters $\nu$ and $m_{\text{stop}}$, B\"uhlmann~et.~al.~\cite{buehlmann07boosting} claim that the \emph{learning rate} $\nu$ is of minor importance as long as it is `sufficiently small', with $\nu = 0.1$ commonly used in practice.

The stopping criterion $m_{\text{stop}}$, determines the degree of regularization and thereby heavily affects the model quality in terms of overfitting and variable selection \cite{mayr2012knowing}.
However, as already outlined in the introduction, optimizing $m_{\text{stop}}$ using common approaches like cross-validation results in the selection of many uninformative variables. 
Although still focusing on minimizing prediction error, using a 25-fold bootstrap instead of the commonly used 10-fold cross-validation tends to return sparser models without sacrificing prediction performance \cite{hepp2016approaches}.

\subsection{Stability Selection}

The weak performance of cross-validation regarding variable selection partly results from the fact that it pursues the goal of minimizing the prediction error instead of selecting only informative variables.
One possible solution is the \emph{stability selection} framework \cite{meinshausen2010stab,shah2013variable}, a very versatile algorithm that can be combined with all kind of variable selection methods like gradient boosting, lasso or forward stepwise selection. It produces sparser solutions by controlling the number of false discoveries.
Stability selection defines an upper bound for the per-family error rate (PFER), e.g., the expected number of uninformative variables $\mathds{E}(V)$ included in the final model.

Therefore, using stability selection with model-based boosting  means that Algorithm~\ref{alg:boosting} is run independently on $B$ random subsamples of the data until either a predefined number of iterations $\ms$ is reached or $q$ different variables have been selected. 
Subsequently, all variables are sorted with respect to their selection frequency in the $B$ sets.
The amount of informative variables is then determined by a user-defined threshold $\pi_\text{thr}$ that has to be exceeded.
A detailed description of these steps is given in Algorithm~\ref{alg:stabsel}.

\begin{algorithm}
\caption{Stability selection for model-based boosting \cite{hofner2014controlling}}
\label{alg:stabsel}
\begin{enumerate}
\item For $b = 1,\dots, B$:
\begin{enumerate}
\item Draw a subset of size $\lfloor n/2\rfloor$ from the data
\item Fit a boosting model to the subset until the number of selected variables is equal to $q$ or the number of iterations reaches a pre-specified number ($m_\text{stop}$).\\
\end{enumerate}
\item Compute the selection frequencies per variable $j$:
\begin{equation}
\label{eq:sel_freq}
\hat\pi_j := \frac{1}{B}\sum^B_{b=1}\mathds{I}_{\{j\in\hat S_{b}\}},
\end{equation}
where $\hat S_{b}$ denotes the set of selected variables in iteration $b$.
\item Select variables with a selection frequency of at least $\pi_\text{thr}$, which yields a set of stable covariates
\begin{equation}
\label{eq:thr}
\hat S_\text{stable} := \{j:\hat\pi_j\ge\pi_\text{thr}\}.
\end{equation}
\end{enumerate}
\end{algorithm}

Following this approach, the upper bound for the PFER can be derived as follows \cite{meinshausen2010stab}:
\begin{equation}
\label{eq:bound}
\mathds{E}(V) \le \frac{q^2}{(2\pi_\text{thr} - 1)p}.
\end{equation}
With additional assumptions on exchangeability and shape restrictions on the distribution of simultaneous selection, even tighter bounds can be derived \cite{shah2013variable}.
While this method is successfully applied in a large number of different applications \cite{haury2012tigress,ryali2012estimation,Thomas2016boosting,mayr2016boosting}, several shortcomings impede the usage in practice:
First of, three additional hyperparameters $\pi_\text{thr}$, PFER and $q$ are introduced.
Although only two of them have to be specified by the user (the third one can be calculated by assuming equality in Equation (\ref{eq:bound}), it is not intuitively clear which parameter should be left out and how to specify the remaining two.
Even though recommendations for reasonable settings for the selection threshold \cite{meinshausen2010stab} or the PFER \cite{hofner2014controlling} are proposed, the effectiveness of these settings is difficult to evaluate in practical settings. 
The second obstacle in the usage of stability selection is the considerable computational power required for calculation. 
Overall $B$ boosting models (\cite{shah2013variable} recommends $B = 100$) have to be fitted and a reasonable $\ms$ has to be found as well, which will most likely require cross-validation. 
Even though this process can be parallelized quite easily, complex model classes with smooth and higher-order effects can become extremely costly to fit.

\subsection{Probing}

The approach of adding \emph{probes} or \emph{shadow variables}, e.g., artificial uninformative variables to the data, is not completely new and has already been investigated in some areas of machine learning.
Although they share the underlying idea to benefit from the presence of variables that are known to be independent from the outcome, the actual implementation of the concept differs (see Guyon and Elisseeff (2003)~\cite{guyon2003introduction} for an overview).
An especially useful approach, however, is to generate these additional variables as randomly shuffled versions of all observed variables.
These permuted variables will be called \emph{shadow variables} for the remainder of this paper and are denoted as $\tilde{x}_j$.
Compared to adding randomly sampled variables, shadow variables have the advantage that the marginal distribution of $x_j$ is preserved in $\tilde{x}_j$. 
This approach is tightly connected to the theory of permutation tests \cite{strasser1999asymptotic} and is used similarly for \emph{all-relevant} variable selection with random forests \cite{kursa2010boruta}. 

Implementing the \emph{probing} concept to the sequential structure of model-based gradient boosting is rather straightforward.
Since boosting algorithms proceed in a greedy fashion and only update the effect which yields the largest loss reduction in each iteration, selecting a shadow variable essentially implies that the best possible improvement at this stage relies on information that is known to be unrelated to the outcome.
As a consequence, variables that are selected in later iterations are most likely correlated to $y$ only by chance as well.
Therefore, all variables that have been added prior to the first shadow variable are assumed to have a true influence on the target variable and should be considered informative.
A description of the full procedure is presented in Algorithm~\ref{alg:probing}.

\begin{algorithm}
\caption{Probing for variable selection in model-based boosting}
\label{alg:probing}
\begin{enumerate}
%

\item \textbf{Expand} the dataset $X$ by creating randomly shuffled images $\tilde{x}_j$ for each of the $j = 1, \dots, p$ variables $x_j$ such that 
\[
\tilde{x}_j \in S_{x_j},
\]
where $S_{x_j}$ denotes the symmetric group that contains all $n!$ possible permutations of $x_j$.

\item \textbf{Initialize} a boosting model on the inflated dataset 
\[
\bar{X}=[x_1 \dots x_p \ \tilde{x}_1 \dots \tilde{x}_p]
\]
and start iterations with $m=0$.


\item \textbf{Stop if} the first $\tilde{x}_j$ is selected, see Algorithm~\ref{alg:boosting} step 3.

\item \textbf{Return} only the variables selected from the original dataset $X$.

\end{enumerate}

\end{algorithm}

The major advantage of this approach compared to variable selection via cross-validation or stability selection is that one model fit is enough to find informative variables and no expensive refitting of the model is required.
Additionally, there is no need for any prespecification like the search space ($m_\text{stop}$) for cross-validation or additional hyperparameters ($q$, $\pi_\text{thr}$, PFER) for stability selection.
However, it should be noted that unlike classical cross-validation, probing aims at optimal variable selection instead of prediction performance of the algorithm.
Since this usually involves stopping much earlier, the effect estimates associated with the selected variables are most likely strongly regularized and might not be optimal for predictions.

%

\section{Simulation study}
\label{sec:simulation}

In order to evaluate the performance of our proposed variable selection method, we conduct a benchmark simulation study where we compare the set of non-zero coefficients determined by the use of shadow variables as stopping criterion to cross-validation and different configurations of stability selection.
We simulate $n$ data points for $p$ variables from a multivariate normal distribution $X\sim\mathcal{N}(0,\Sigma)$ with Toeplitz correlation structure $\Sigma_{ij}=\rho^{|i-j|}$ for all $1<i,j<p$ and $\rho=0.9$.
The response variable $y^{(i)}$ is then generated by sampling Bernoulli experiments with probability
$$ \pi^{(i)} = \frac{\exp(\eta^{(i)})}{1+\exp(\eta^{(i)})},$$
with $\eta^{(i)}$ the linear predictor for the $i$th observation $\eta^{(i)} = X^{(i)}\beta$ and all non-zero elements of $\beta$ sampled from $\mathcal{U}(-1,1)$.
Since the total amount of non-zero coefficients determines the number of informative variables in the setting, it is denoted as $p_\text{inf}$.

Overall, we consider 12 different simulation scenarios defined by all possible combinations of $n \in \{100, 500\}$, $p \in \{100, 500, 1000\}$ and $p_\text{inf} \in \{5, 20\}$.
Specifically, this leads to the evaluation of 2 low-dimensional settings with $p<n$, 4 settings with $p=n$ and 6 high-dimensional settings with $p>n$.
Each configuration is run $100$ times.
Along with new realizations of $X$ and $y$, we also draw new values for the non-zero coefficients in $\beta$ and sample their position in the vector in each run to allow for varying correlation patterns among the informative variables.
For variable selection with cross-validation, $25$-fold bootstrap (the default in \texttt{mboost}) is used to determine the final number of iterations. 
Different configurations of stability selection were tested to investigate whether and, if so, to what extent these settings affect the selection.
In order to explicitly use the upper error bounds of stability selection, we decided to specify $9$ combinations with $\text{PFER}\in\{1, 2.5, 8\}$ and $\pi_\text{thr}\in\{0.6, 0.75,0.9\}$ and calculate $q$ from Equation (\ref{eq:bound}).
Aside from the learning rate $\nu$, which is set to $0.1$ for all methods, no further parameters have to be specified for the probing scheme.
Two performance measures are considered for the evaluation of the methods with respect to variable selection:
First, the true positive rate (TPR) as the fraction of (correctly) selected variables from all true informative variables and second, the false discovery rate (FDR) as the fraction of uninformative variables in the set of selected variables.
To ensure reproducibility the R package \texttt{batchtools}~\cite{pkg_batchtools:2015} was used for all simulations.
The data and code to fully reproduce the simulation study can be found in the online supplementary materials of this manuscript.

The results of the simulations for all settings are illustrated in Figure~\ref{fig:bivar}.
With TPR and FDR on the y-axis and x-axis, respectively, solutions displayed in the top left corner of the plots therefore successfully separate the $p_\text{inf}$ informative variables from the ones without true effect on the response.
Although already using a sparse cross-validation approach, the FDR of variable selection via cross-validation is still relatively high, with more than 50\% false positives in the selected sets in the majority of the simulated scenarios.
Whereas this seems to be mostly disadvantageous in the cases where $p_\text{inf}=5$, the trend to more greedy solutions leads to a considerably higher chance of identifying more of the truly informative variables if $p_\text{inf}=20$ or with very high $p$, however still at the price of picking up many noise variables on the way.
Pooling the results of all configurations considered for stability selection, the results cover a large area of the performance space in Figure~\ref{fig:bivar}, thereby probably indicating high sensitivity on the decisions regarding the three tuning parameters.

\begin{sidewaysfigure}
\centering
\includegraphics{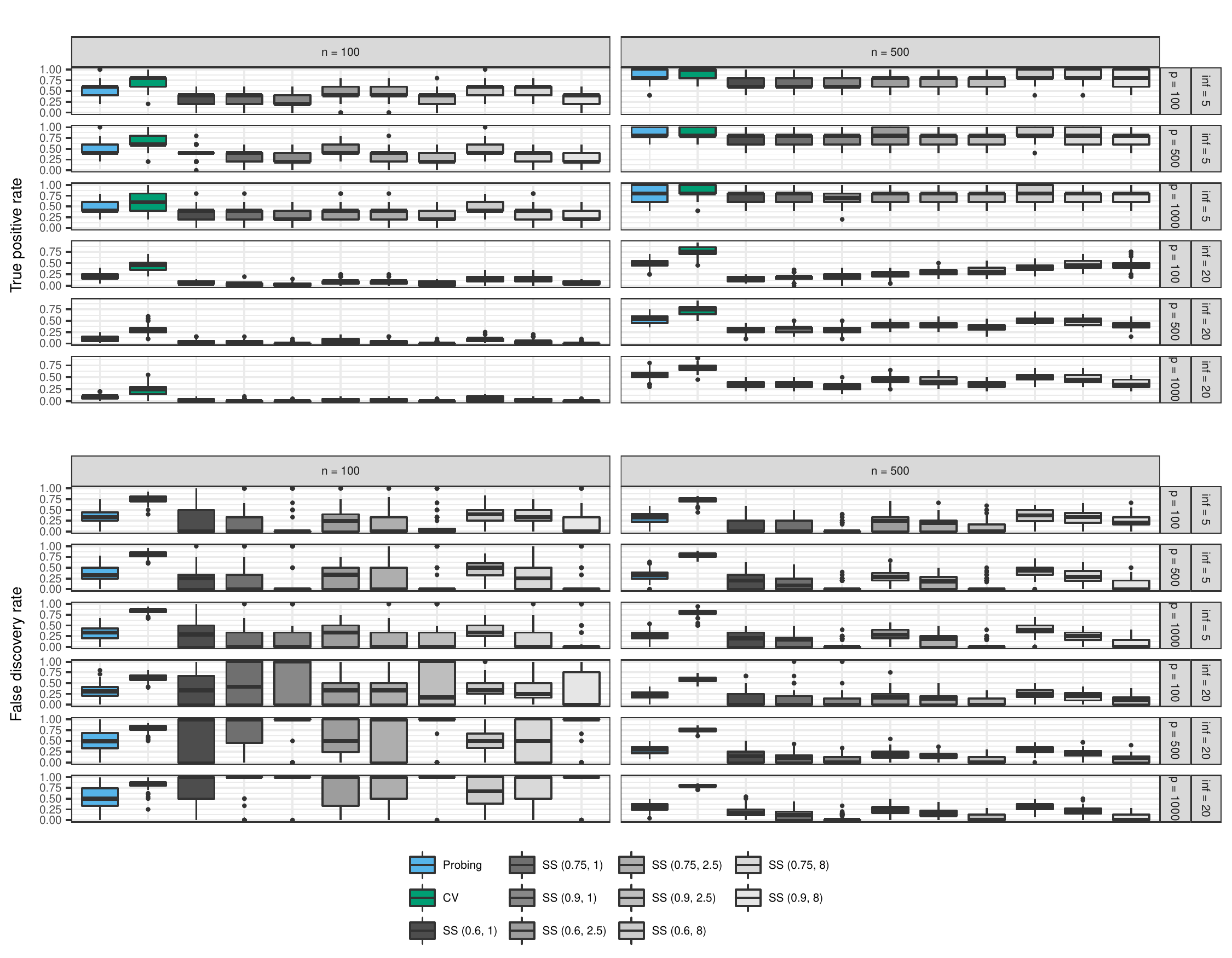}
\caption{Boxplots of true positive rate (top) and false discovery rate (bottom) for different simulation settings and the three boosting-based, variable selection algorithms. Different Stability selection settings are denoted by $SS(\pi_\text{thr}, \text{PFER})$}
\label{univar}
\end{sidewaysfigure}

Examining the results separately in Figure~\ref{univar}, the dilemma is particularly clearly illustrated for $p_\text{inf}=20$ and $n=500$.
Although being able to control the upper bounds for expected false positive selections, only a minority of the true effects are selected if the PFER is set too conservative.
In addition, the high variance of the FDR observed for these configurations in some settings somewhat counteracts the goal to achieve more certainty about the selected variables one might probably pursue by setting the PFER very low.
The performance of probing, on the other hand, reveals a much more stable pattern and outperforms stability selection in the difficult $p_\text{inf}=20$ and $n=100$ settings.
In fact, the TPR is either higher or similar to all configurations used for stability selection, but exhibiting slightly higher FDR especially in settings with $n=500$.
Interestingly, probing seems to provide results similar to those of stability selection with PFER=8, raising the question if the use of shadow variables allows statements about the number of expected false positives in the selected variable set.

Considering the runtime, however, we can see that probing is orders of magnitudes faster with an average runtime of less than a second compared to 12 seconds for cross-validation and almost one minute for stability selection.

\begin{knitrout}
\definecolor{shadecolor}{rgb}{0.969, 0.969, 0.969}\color{fgcolor}\begin{figure}
\includegraphics[width=\maxwidth]{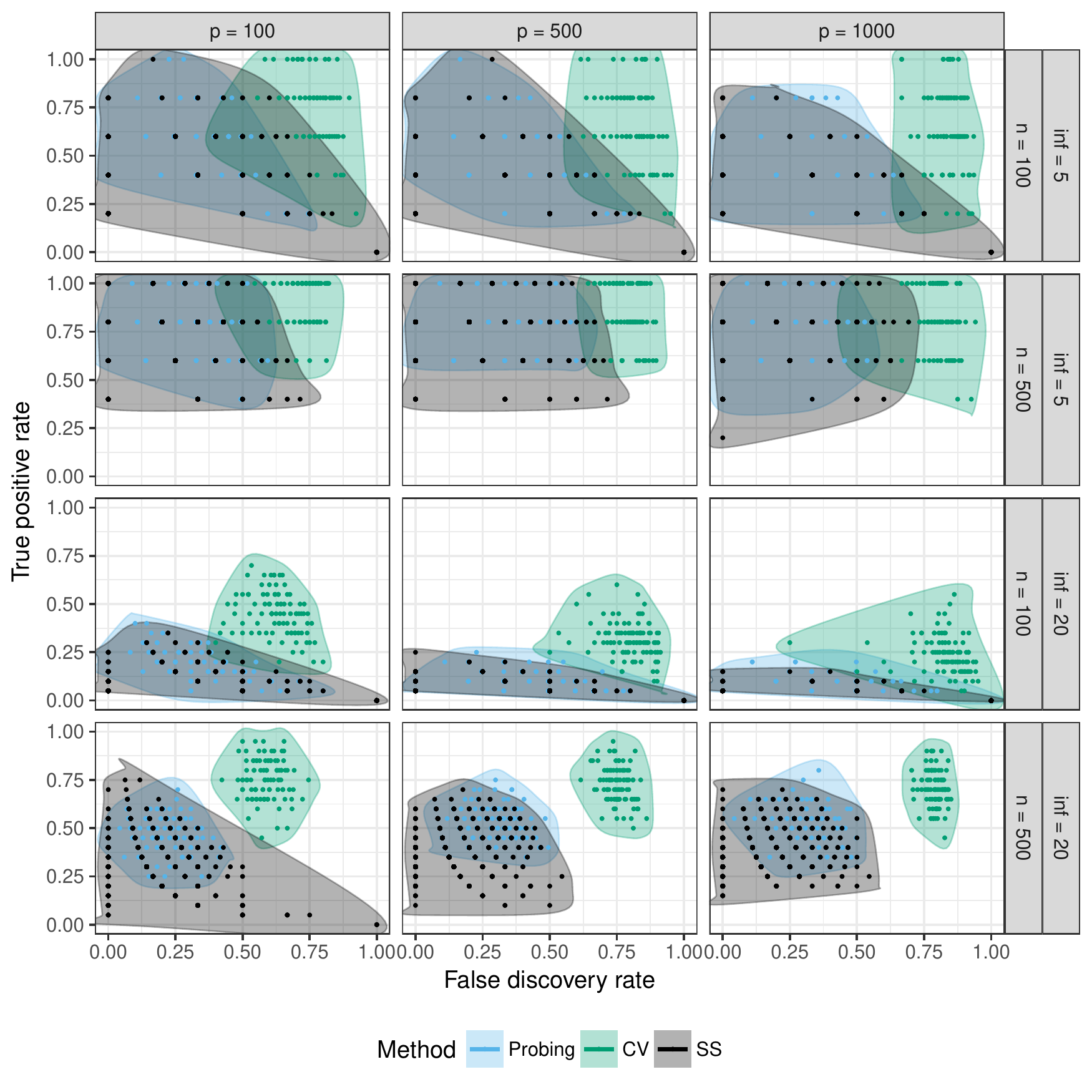} \caption[True positive rate (on y-axis) and false discovery rate (on x-axis) for three different, boosting-based variable selection algorithms, probing (black), stability selection (green), cross-validation (blue) and diffierent simulation settings]{True positive rate (on y-axis) and false discovery rate (on x-axis) for three different, boosting-based variable selection algorithms, probing (black), stability selection (green), cross-validation (blue) and diffierent simulation settings: $n \in \{100, 500\}$, $p \in \{100, 500, 1000\}$ and $p_{\text{inf}} \in \{ 5, 20\}$. All settings of stability selection are combined. Shaded areas are smooth hulls around all observed values.}\label{fig:bivar}
\end{figure}

\end{knitrout}

%

\section{Gene expression data}
\label{sec:application}

In this Section we exploit the usage of probing as a tool for variable selection on a gene expression data set.
The data set examines riboflavin production by Bacillus subtilis~\cite{buhlmann2014high} with $n=71$ observations of log-transformed riboflavin production rates and expression level for $p=4088$ genes and is publicly available in the \texttt{R} package \texttt{hdi}.
Our proposed probing approach is implemented in a fork of the \texttt{mboost}~\cite{mboost2016} software for component-wise gradient boosting. 
It can be easily used by setting \texttt{probe=TRUE} in the \texttt{glmboost()} call.


In order to evaluate the results provided by the new approach, we analysed the data with cross-validation (the \texttt{mboost} default 25-fold bootstrap), stability selection~\cite{Hofner:pkg_stabs:2015} and the lasso~\cite{glmnet2010} for comparison. 
Table~\ref{tab:sels} shows the total number of variables selected by each method along with the size of the intersection between the sets.
Starting with the probably least surprising result, boosting with cross-validation leads to a very large set of $50$ selected variables, while using probing as stopping criterion instead reduces the set to contain only 10 variables.
Since both approaches are based on the same regularization profile until the first shadow variable enters the model, the less regularized solution of cross-validation contains all these 10 variables as well.
For stability selection, we adopted the configuration of $\text{PFER}=1$ and $q = 20$ used in B\"uhlmann et. al. (2014) \cite{buhlmann2014high}.
As a consequence, the set of variables deemed to be informative shrinks to only 5.
These results clearly reflect the findings from the simulation study in Section \ref{sec:simulation}, placing the probing approach between stability selection with probably overly conservative error bound and the greedy selection with cross-validation.

Since so far all approaches rely on boosting algorithms, we additionally considered variable selection with the lasso.
We used the default settings of the \texttt{glmnet} package for R to calculate the lasso regularization path and determine the final model via 10-fold cross-validation \cite{glmnet2010}.
Although the overall model size of 30 variables is much higher than the result for Probing, they only agree on 7 mutually selected variables.
Interestingly, even one of the 5 variables proposed by stability selection is also missing.
The \texttt{R} code used for this analysis can be found in the online supplementary materials of this manuscript. 

\begin{table}
\centering
\begin{tabular}{rcccc}
\toprule
&\multicolumn{4}{c}{Selected variables}\\
\cmidrule(r){2-5}
    & Cross-Val. & Probing  & StabSel & Glmnet\\
\midrule
Cross-Val.& $50$  &       &     &       \\
Probing   & $10$  & $10$  &     &       \\
StabSel   & $5$   & $5$   & $5$ &       \\
Glmnet    & $23$  & $7$   & $4$ & $30$  \\
\bottomrule
\end{tabular}
\caption{Number of selected variables (on diagonal) and number of identical selected variables by two methods (on off-diagonal) for four variable selection techniques (boosting with 25-fold bootstrap, probing, stability slection and the lasso estimated via \texttt{glmnet}) on Ribovlafin Data~\cite{buhlmann2014high}.}
\label{tab:sels}
\end{table}

\section{Conclusion}
\label{sec:conclusion}

We proposed a new approach to determine the optimal number of iterations for sparse and fast variable selection with model-based boosting based on adding probes or shadow variables (\textit{probing}).
We were able to demonstrate via a simulation study and the analysis of gene expression data that our approach is both a feasible and convenient strategy for variable selection in high-dimensional settings.
In contrast to common tuning procedures for model based boosting which rely on resampling or cross-validation procedures to optimize the prediction accuracy \cite{mayr2012knowing}, our probing approach directly addresses the variable selection properties of the algorithm.
As a result, it substantially reduces the high number of false discoveries that arise with standard procedures \cite{hofner2014controlling} while only requiring a single model fit to obtain the set of parameters.

Aside from the very short runtime, another attractive feature of probing is that no additional tuning parameters have to be specified to run the algorithm.
While this greatly increases its ease of use, there is, of course, a trade-off with respect to flexibility, as the lack of tuning parameters means that there is no way to steer the results towards more or less conservative solutions.
However, a corresponding tuning approach in the context of probing could be to allow a certain amount of selected probes in the model before deciding to stop the algorithm (cf. Guyon et al., 2003 \cite{guyon2003introduction}).
Although variables selected after the first probe can be labeled informative less convincingly, this resembles the uncertainty that comes with specifying higher values for the error bound of stability selection.

A potential drawback of our approach is that due to the stochasticity of the permutations, there is no deterministic solution and the selected set might slightly vary after rerunning the algorithm.
In order to stabilize results, probing could also be used combined with resampling to determine the optimal stopping iteration for the algorithm by running the procedure on several bootstrap samples first.
Of course, this requires the computation of multiple models and therefore again increases the runtime of the whole selection procedure.

Another promising extension could be a combination with stability selection. With each model stopping at the first shadow variable, only the selection threshold $\pi_\text{thr}$ has to be specified.
However, since this means a fundamental change of the original procedure, further research on this topic is necessary to better assess how this could affect the resulting error bound. 

While in this work we focused on gradient boosting for binary and continuous data, there is no reason why our results should not also carry over to other regression settings or related statistical boosting algorithms as likelihood-base boosting \cite{TutzBinder}.
Likelihood-based boosting follows the same principle idea but uses different updates, coinciding with gradient boosting in case of Gaussian responses \cite{review2014evolution}.
Further research is also warranted on extending our approach to multidimensional boosting algorithms \cite{Thomas2016boosting,gamboostlss:2012}, where variables have to be selected for various models simultaneously. 

In addition, probing as a tuning scheme could be generally also combined with similar regularized regression approaches like the lasso \cite{tibshirani96lasso,hepp2016approaches}.
Our proposal for model-based boosting hence could be a starting point for a new way of tuning algorithmic models for high-dimensional data -- not with the focus on prediction accuracy, but addressing directly the desired variable selection properties. 

\subsection*{Acknowledgements and disclosure}
The work of authors TH and AM was supported by the Interdisciplinary Center for Clinical
Research (IZKF) of the Friedrich-Alexander-University Erlangen-N\"urnberg (Project J49).\\

The authors declare that there is no conflict of interest regarding the publication of this paper.


\bibliographystyle{abbrv}
\bibliography{literatur}
\end{document}